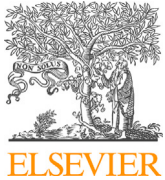

Contents lists available at ScienceDirect

# Information Sciences

journal homepage: www.elsevier.com/locate/ins

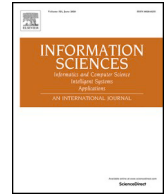

# SDC-HSDD-NDSA: Structure detecting cluster by hierarchical secondary directed differential with normalized density and self-adaption

Hao Shu

*Shenzhen University, Shenzhen, 518060, Guangdong, PR China*

A R T I C L E  I N F O

A B S T R A C T

Density-based clustering is the most popular clustering algorithm since it can identify clusters of arbitrary shape as long as they are separated by low-density regions. However, a high-density region that is not separated by low-density ones might also have different structures belonging to multiple clusters. As far as we know, all previous density-based clustering algorithms fail to detect such structures. In this paper, we provide a novel density-based clustering scheme to address this problem. It is the first clustering algorithm that can detect meticulous structures in a high-density region that is not separated by low-density ones and thus extends the range of applications of clustering. The algorithm employs secondary directed differential, hierarchy, normalized density, as well as the self-adaption coefficient, called Structure Detecting Cluster by Hierarchical Secondary Directed Differential with Normalized Density and Self-Adaption, dubbed SDC-HSDD-NDSA. Experiments on synthetic and real datasets are implemented to verify the effectiveness, robustness, and granularity independence of the algorithm, and the scheme is compared to unsupervised schemes in the Python package *Scikit-learn*. Results demonstrate that our algorithm outperforms previous ones in many situations, especially significantly when clusters have regular internal structures. For example, averaging over the eight noiseless synthetic datasets with structures employing ARI and NMI criteria, previous algorithms obtain scores below 0.6 and 0.7, while the presented algorithm obtains scores higher than 0.9 and 0.95, respectively.[1]

## 1. Introduction

The significance of machine learning has been increasingly recognized in recent years with substantial improvements in different directions such as natural language processing (NLP) [1,2], computer vision(CV) [3], with wide applications such as in energy-efficiency improving [4,5], ergonomics risk assessment [6], curved text detection [7], price-forecasting [8,9]. The study of low-level tasks such as Edge detection [10] and clustering [11,12], especially with unsupervised methods, is of great importance to profoundly understand the dynamics of machine learning, since such low-level tasks are the most fundamental ones while unsupervised methods are more explainable than supervised ones. In this paper, one of the most classical and fundamental low-level tasks in unsupervised machine learning, clustering, is studied.

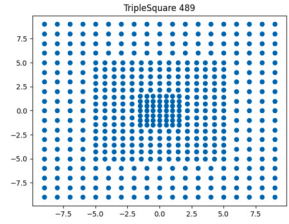

**Fig. 1. (Triple-Square.)** The region consists of three squares, the inner square with the highest density, the middle region with the middle density, and the outer region with the lowest density. Previous density-based algorithms can only detect a unique cluster on it depending on the density threshold, i.e., they cannot distinguish the three regions without missing.

For example, the DBSCAN [12] algorithm could only detect a unique cluster no matter how the density threshold is set. It will divide all the regions into a single cluster with a very-low-density threshold, miss the outer region and divide the inner two regions into a single cluster with a middle-density threshold, or only detect the inner square as a single cluster while missing both the middle and the outer ones with a higher-density threshold.

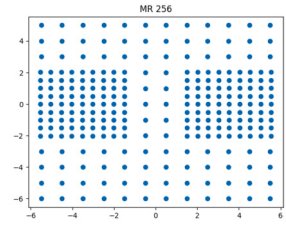

**Fig. 2. (Mountain-River.)** The dataset consists of three regions, two of which have higher density and one of which has a lower density. It looks like a river crosses two mountains. Previous density-based algorithms cannot detect all three clusters, namely they either predict a unique cluster with a density threshold lower than the low-density region or predict two clusters with a density threshold higher than the low-density region but lower than the high-density regions.

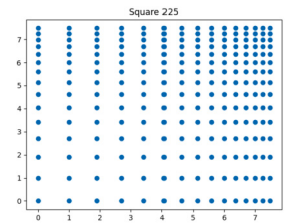

**Fig. 3. (Square.)** The region is a square with gradually changed density, consisting of a unique cluster. The density of the top-right can be extremely high while the density of the bottom-left can be extremely low. When it is put together with other regions (such as in Fig. 5), previous density-based algorithms can either merge all regions into a single cluster with an extremely low-density (lower than all densities in the whole dataset) threshold or have a miss of the bottom-left in the cluster containing the region (with a not extreme low-density threshold).

Clustering algorithms aim to group data by certain criteria and have been investigated for decades. They have extensive applications in multiple subjects such as data mining [13] and astronomical classification [14]. Currently, density-based clustering algorithms could be the most outstanding ones among all clustering algorithms for their ability to identify clusters in different shapes and robustness against noises. However, previous dense-based clustering algorithms are built on the same preconditions that any single cluster should have a high density while different clusters are separated by low-density regions, while, a high-density region might have internal structures that are not separated by low-density regions. For example, in Fig. 1, Fig. 2, and Fig. 3, the region obviously has 3, 3, and 1 clusters, respectively. However if one implements previous density-based algorithms on them, the clustering result could be counter-intuitive. The DBSCAN [12] algorithm could detect a unique cluster in the case of Fig. 1 even with a suitable density threshold. It will divide all the regions into a single cluster with a very-low-density threshold, miss the outer region and divide the inner two regions into a single cluster with a middle-density threshold, or only detect the inner square as a single cluster while missing both the middle and the outer ones with a higher-density threshold. The reason why it fails is that there could be structures in a high-density region, which are not separated by low-density regions but belong to different clusters, and thus the precondition of DBSCAN algorithm is not satisfied. Indeed, as far as we know, all previous algorithms fail on such datasets.

To solve this problem, in this paper, we present a novel clustering scheme that can not only detect clusters separated by low-density regions but also detect different structures in high-density regions even though they are not separated by low-density regions. As it removes the precondition that clusters should be separated by low-density regions, which is required by most of the previous algorithms, our algorithm extends the range of applications of clustering. Moreover, the scheme is robust over noises and independent of the granularities of data. It employs secondary directed differential, hierarchical method, normalized density, as well as the self-adaption coefficient, and thus named Structure Detecting Cluster by Hierarchical Secondary Directed Differential with Normalized Density and Self-Adaption, abbreviated SDC-HSDD-NDSA for short. Also, we provide experiments on several datasets in different granularities, with or without structures or noises, and including synthetic ones and real ones, to demonstrate its validity and robustness.

The organization of the paper is as follows. In Section 2, Previous clustering algorithms and preliminary knowledge are reviewed. Then, Section 3 is dedicated to presenting the main scheme, followed by the experimental results in Section 4. Some discussions, including the study of complexities and the illustration of limitations, are provided in Section 5, and Section 6 is a conclusion section.

The main contributions of the work include:

(1) Illustrate the limitations of previous clustering algorithms that all fail to detect structures in high-density regions and provide representative datasets containing clusters with internal structures.

(2) Provide a novel clustering scheme, dubbed SDC-HSDD-NDSA, which not only enjoys the ability that previous algorithms have but is also valid in detecting structures in high-density regions, and thus extends the range of application of clustering.

(3) Provide experiments on synthetic and real datasets to demonstrate the effectiveness, robustness, and granularity independence of the new method.

The main insights of our scheme include:

(1) To detect structures in a high-density region that is not separated by low-density ones, employing density as the criterion is insufficient. Indeed, even employing the differential of density could be insufficient. Please see the types in Fig. 1 in which the differential of density could be lower but should have more than one cluster and Fig. 3 in which the differential of density could be significantly higher but should have only one cluster.[2]

(2) When processing data of dimensions more than 1, the differentials are essentially directed, and thus employing non-directed properties such as the gradient value can be insufficient. Please see Fig. 2 in which there should be three clusters. The two columns in the middle should be in the same cluster of the rows at the top and down, but the gradient value (might be defined as the maximal differential of a point from points in its neighbor) of points in middle columns could be significantly different from the top and down rows.

(3) If we employ $k$ nearest neighbors, choosing different $k$ in calculating densities and searching closed points might improve algorithms.

(4) To avoid the influence of granularity, employing normalized density should be beneficial. For example, it might allow coefficients independent of datasets. On the other hand, it might be needed in different parts of an algorithm to choose different normalized schemes.

(5) A normalization scheme alone is insufficient for obtaining granularity independence since it can only transform the granularity of a dataset into a given range but cannot process datasets consisting of clusters with extreme granularity differences. Therefore, to obtain granularity independence by normalizing schemes, other techniques such as hierarchy schemes should be applied to make the dataset (or subdatasets) consist of clusters with similar granularities.

(6) The self-adaption of the coefficient could be obtained by repeating the algorithm several times.

(7) Various strategies might be needed in different parts of an algorithm.

## 2. Review

### 2.1. A brief review of clustering algorithms

A satisfactory clustering algorithm should group data without violating intuition, be robust over noises, and also be expected to be valid in wide granularities with low complexity. In the past few decades, many algorithms have been proposed and can be divided into several categories such as partition-based ones [11,15,16] which view the cluster task as a partition task, model-based ones [17,18] which cluster data by some fixed models, density-based ones [12,19–24] which employ density as the cluster criterion, grid-based ones [25,26] which advocate grid structures for clustering, border-based ones [27–29] which try to determine clusters by their borders, and hierarchical ones [23,30–35] which gradually merge or split clusters to obtain the final result. Moreover, similar ideas are employed in other tasks such as detecting isolated points [36,37]. Also, there are several papers focusing on accelerating the algorithms [38–43].

The most famous density-based clustering algorithm was proposed in 1996, known as DBSCAN [12]. It marks points with a sufficiently high density as core ones, and then groups core points into the same cluster with its neighbors. Most later works could somehow be traced back to it. There are many varieties of density-clustering algorithms, such as DENCLUE [44,45] which calculates the influence of each point first, OPTICS [19] which orders the points first, LOF [36] which employs the $k$-nearest neighbor to detect outliers, HDBSCAN [33] which employs hierarchical algorithms in DBSCAN, DP [21,24] which employs density peaks, ADBSCAN [22] which employs the adapting coefficient, and algorithms employing shared nearest neighbor [46,47] or reversed nearest neighbor [48–50] to define density.

Despite the advantages of density-based algorithms that can cluster data with multiple kinds of shapes, previous density-based clustering algorithms are built on the same preconditions that single clusters should be separated by low-density regions. Therefore, they would fail if a high-density region has internal structures belonging to different clusters.

---

[2] Consider the case that a dataset consists of both the type in Fig. 1 and the type in Fig. 3 which are separated by low-density regions, for example, the dataset of Figure 18 in the supplemental material. In such a case, a region with a higher differential of density might only have 1 cluster while a region with a lower differential of density might have more than 1 cluster.

## 2.2. Preliminary knowledge

This subsection aims to review the DBSCAN algorithm and its $k$-NN version, as well as provide related knowledge for understanding the scheme in the next section. We will explain the definitions and algorithms as simply as possible. As they might be textbook level, readers familiar with this subject might skip.

**Definition 1** (*ϵ-neighbor*). Let $X$ be a dataset, $x \in X$ a point, and $\epsilon \geq 0$ a non-negative real number. The $\epsilon$-neighbor $N_\epsilon(x)$ of $x$ is the set of points in $X$ whose distance to $x$ is at most $\epsilon$:

$$N_\epsilon(x) := \{y \in X \mid D(x, y) \leq \epsilon\},$$

where $D(x, y)$ is a fixed distance metric.

In the DBSCAN algorithm [12], the density of a point $x$ is defined to be the number of points in its $\epsilon$-neighbor, namely $|N_\epsilon(x)|$. To run the algorithm, one must choose an $\epsilon$ and a minimal density $Minst$. Points whose density is larger than $Minst$ are marked as core points. The algorithm is implemented as follows. Firstly, all points are marked as unclassified, while an unclassified core point is picked to form a cluster. Then the cluster is expanded iteratively by absorbing the unclassified points in the $\epsilon$ neighbors of the core points into the same cluster−−when a new core point is absorbed, the $\epsilon$ neighbor of it is also absorbed. The absorbed points are marked as classified, and the cluster is expanded until no new points can be absorbed. Hence, a final cluster is defined. After that, a new unclassified core point is picked to form a new cluster and is expanded as above. The procedure is repeated until no core points are unclassified. Finally, all left unclassified points are marked as noises, while the above-form clusters consist the final clustering result.

Although the DBSCAN algorithm has caught substantial attention since it was published, one of its main flaws is the need for choosing $\epsilon$ and $Minst$, which might be challenging to decide since it requires prior knowledge such as the granularity of the data set. By enlarging the granularity of a data set, for example, changing the coordinates of all data $x = (x_1, x_2)$ in a two-dimensional space into $nx = (nx_1, nx_2)$, the clustering result should not be changed. Hence, $\epsilon$ should be substituted with another, namely $n\epsilon$. Even if this issue could be solved by data preprocessing, the difficulty in choosing $\epsilon$ and $Minst$ might still occur, since different clusters might have different local granularities.

To address the issue in choosing $\epsilon$, a suggestion might be employing the $k$-nearest neighbor instead of the $\epsilon$-neighbor.

**Definition 2** (*k-Nearest Neighbors (k-NN)*). Let $X$ be a dataset, $x \in X$ a point, and $k \geq 1$ a positive integer. A $k$-distance point $p_k(x) \in X$ for $x$ satisfies the following conditions:

- There exist at most $k-1$ points $y \in X$ such that $D(x, y) < D(x, p_k(x))$.
- There exist at least $k$ points $y \in X$ such that $D(x, y) \leq D(x, p_k(x))$.

The $k$-nearest neighbors ($k$-NN) of $x$ are defined as the set of points in $X$ whose distance to $x$ does not exceed $D(x, p_k(x))$, namely $\{y \in X \mid D(x, y) \leq D(x, p_k(x))\}$.

Simply speaking, a $k$-distance point of $x$ is a $k$-th closest point of $x$ (there might be several points with the same distances to $x$) while the $k$-NN of $x$ consists of points not farther than a $k$-distance point.

The $k$-nearest neighbor was first employed to detect isolated points [36], but it could be easily employed in clustering tasks. It can be viewed as a method for choosing $\epsilon$ in the DBSCAN algorithm−−different $\epsilon$ might be chosen for different points to be the $k$-distances of them. The $k$-NN version of the DBSCAN algorithm can be obtained simply by replacing the $\epsilon$-neighbor with the $k$-NN of points with a prefixed $k$ after marking all isolated points and viewing all remaining ones as core points.[3]

Nevertheless, to detect isolated points, criteria like $Minst$ in the DBSCAN algorithm, which depends on the granularity, must be chosen−−for example, the maximal $k$-distance for a point not marked as an isolated point.

## 3. Methodology

Algorithms, illustrations, and analyses are presented in this section. The pseudocode of the core of the SDC-HSDD-NDSA algorithm is provided in subsection 3.1, followed by illustrations in subsections 3.2 to 3.5. Some drawbacks of the core algorithm are investigated in subsection 3.6, and to overcome them, the SDC-HSDD-ND algorithm is proposed in subsection 3.7. Subsection 3.8 is dedicated to a suggestion for choosing coefficients, and the final algorithm with the self-adaption coefficient, namely SDC-HSDD-NDSA algorithm, is presented in subsection 3.9. The Python codes of the pseudocodes can be found on https://github.com/Hao-B-Shu/SDC-HSDD-NDSA.

---

[3] The above algorithm will be called the $k$-NN clustering algorithm in the remaining paper, while the algorithm predicting each new point to be in the cluster containing the closest known point will be named the $k$-NN classifier.

### 3.1. The core algorithm: SDC-SDD-ND

In summary, the core of the SDC-HSDD-NDSA algorithm, dubbed SDC-SDD-ND is implemented by the following steps.

Step 1 (Lines 1-6): Calculate the density of points and normalize the density.

Step 2 (Line 7): Calculate the differential of point $a$ and point $b$ for all pairs of points $(a,b)$ such that $b$ is in the neighborhood of $a$.

Step 3 (Line 8): Mark points with normalized density lower than a threshold as isolated ones, forming a special set.

Step 4 (Lines 9-34): Cluster points based on the directed secondary differential of points, namely if $a$ is in a cluster and $b$ is a neighbor of $a$, then add $b$ into the same cluster if there exists a neighbor $e$ of $b$ with the differential of density of $a,b$ minus the differential of density of $b,e$ being small enough.

Step 5 (Line 35): Merge points in a cluster that is too small with the closest one.

In detail, the pseudocode for SDC-SDD-ND is in Algorithm 1, where only the dataset is required and other inputs are optional with defaults discussed in the following subsections.

### 3.2. The calculation of the densities

The first task needed in all density-based algorithms is to calculate the densities. Some papers employed $\frac{1}{\epsilon}$, where $\epsilon$ is the $k$-distance of the point, to be the density of a point. However, we find it more suitable to employ the average distance and thus use $\frac{1}{r^d}$ as the density of a point in the algorithm,[4] where $r$ is the average distance of the closest $RhoCalculateK$ points to the point and $d$ is the dimension of the data.

The coefficient $RhoCalculateK$ is expected to be small. For instance, in Fig. 1, the points in the corners of a square should be considered to have the same density as points in the interior of the square. If $RhoCalculateK$ is chosen to be 2 or 3, it is certainly such a case, while if $RhoCalculateK$ is chosen to be larger than 3, the density of the corner points could be lower than that of the interior points. However, the smaller $RhoCalculateK$ is, the less robust the algorithm becomes since the density of a point could be influenced more easily by a single noisy point. Therefore, considering the trade-offs of these issues, the default value of $RhoCalculateK$ is set to 4 in the algorithm.

On the other hand, to avoid the influence of granularity, it is essential to use normalized densities instead of densities. Furthermore, it could be better to normalize the densities used in calculating secondary differentials and detecting isolated points by different methods. In the algorithm, the densities used in calculating secondary differentials are normalized by dividing by the maximum, while the densities used in detecting isolated points are normalized by dividing by the average.

### 3.3. The searching neighbors

The condition that the secondary differential is lower than $eps$ is very tight, which might be necessary for detecting structures but could also result in the over-refinement of the clusters. To mitigate the issue, other conditions should be relaxed. Therefore, in the algorithm, we suggest increasing the number of searching neighbors. However, using a larger number of neighbors in calculating densities might lead to inaccuracy, as explained in the previous subsection. As a solution, different numbers of neighbors in neighbor searching and density calculating are advocated. Also, a similar idea could be employed in calculating densities in detecting isolated points.

Hence, the presented algorithm uses the number of searching neighbors $SearchNeiborK = 7$, the number of neighbors for calculating densities $RhoCalculateK = 4$, and the number of neighbors for calculating densities employed in detecting isolated points $IsoNeiborK = 4$ as defaults.

### 3.4. The choice of $MaxIsoPointRho$ as the threshold of isolated points

The choice of $MaxIsoPointRho$, namely the lowest density for a point not to be considered isolated, depends on the tightness of detecting isolated points. It should be chosen larger if one requires a stricter standard in detecting isolated points, while it could be lower if only too extreme points are needed to be marked as isolated points. Additionally, $MaxIsoPointRho$ can be set to 0 if detecting isolated points is not needed. In the algorithm above, the default is set to 0.07, based on test results.

### 3.5. The merging of the clusters

As illustrated in Subsection 3.3, the tightness of the secondary differential condition might lead to the over-refinement of the clusters. Consequently, some clusters that should not be separated might still be separated during clustering. However, fortunately, due to the same tightness of the condition, the over-refined clusters are often small in most cases. Therefore, the problem can be managed by merging small clusters.

Regarding the merging scheme, the algorithm simply redistributes points from clusters whose size is smaller than $MinClusterPoint$ to the nearest cluster whose size reaches $MinClusterPoint$. It may not be the best choice, but it is the simplest one and sufficient

---

[4] Employing $\frac{1}{r^d}$ rather than $\frac{1}{r}$ is because it is more similar to the natural density.

**Algorithm 1** The core algorithm: SDC-SDD-ND.

**Input:** *Data*
    %(Required) Data set
**Input:** *MaxIsoPointRho*
    %(Optional) Minimal not isolated density
**Input:** *MinClusterPoint*
    %(Optional) Minimal points required in a cluster
**Input:** *Datanam*
    %(Optional) Name of the Dataset
**Input:** *DataClustername*
    %(Optional) Name of the output result
**Input:** *SearchNeiborK*
    %(Optional) $k$ of $k$-NN
**Input:** *RhoCalculateK*
    %(Optional) Points employed to calculate density
**Input:** *IsoNeiborK*
    %(Optional) Point employed in calculating density when detecting isolated points
**Input:** *eps*
    %(Optional) Absorbing criteria in expanding clusters
**Output:** Cluster result

1: $kNNdis = k$-NN Distance matrix of points
    % $k = max(SearchNeiborK, RhoCalculateK, IsoNeiborK)$
2: $kNNpoint = k$-NN of points
    % $k = max(SearchNeiborK, RhoCalculateK, IsoNeiborK)$
3: $kNrho =$ Density of points calculated by their $RhoCalculateK$ nearest points
4: $Nrho =$ Normalized $kNrho$
5: $iskNrho =$ Isolated density of points calculated by their $IsoNeiborK$ nearest points
6: $Nisrho =$ Normalized $iskNrho$
7: $Drho =$ Differential matrix of points, calculated by $Nisrho$
    %Only calculate for $SearchNeiborK$-NN points of each point
8: $IP =$ The set of all isolated points
    %Points with $iskNrho < MinClusterPoint$
9: $UCNumber =$ The number of non-clustered points
    % Initially, $UCNumber =$ the number of data$-|IP|$
10: $CP$
    % $CP[i] == x$ represents data $i$ in the $x$-th cluster
    % $CP[i] == -1$ means data $i$ has not been clustered
    %All $CP[i] = -1$, initially
11: $C =$ The list of all clusters, initially consisting of $IP$ only
12: Set $CP[i] = 0$ for all points in $IP$
13: $ClusterNumber = 0$
    %The number of clusters
14: **while** $UCNumber > 0$ **do**
15:     $TC =$ A new cluster, initially consisting of the data $i$ with the highest density such that $CP[i] == -1$
16:     $ClusterNmber = ClusterNumber + 1$
17:     $Tem =$ The seed to expand $TC$
        %Initially consisting of the point in $TC$
18:     $UCNumber = UCNumber - 1$
19:     $CP[i] = ClusterNumber$
20:     **while** $Tem \mathrel{!}= \emptyset$ **do**
21:         $a = Tem[0]$
22:         **for** each $b \in kNNpoint[a]$ with $CP[b] == -1$ **do**
23:             **for** each $e \in kNNpoint[b]$ with $CP[e] != 0$ **do**
24:                 **if** abs(Drho[$a$][$b$]-Drho[$b$][$e$]) $< eps$ **then**
25:                     add $b$ into $TC$ and $Tem$
26:                     $CP[b] = ClusterNumber$
27:                     $UCNumber = UCNumber - 1$
28:                 **end if**
29:             **end for**
30:         **end for**
31:         $Tem$ removes $a$
32:     **end while**
33:     Add $TC$ to $C$
34: **end while**
35: **Merge:** Cancel clusters whose length is less than $MinClusterPoint$ and add their points into the closest cluster whose length is not less than $MinClusterPoint$ except $IP$. If the lengths of all non-$IP$ clusters are less than $MinClusterPoint$, then all points not in $IP$ should be merged to the same cluster.
36: Return the final clustering result

for most cases. Other merging methods could also be applied, such as redistributing points from small clusters to the nearest cluster without considering its size, merging clusters instead of redistributing individual points, or using the $k$-NN algorithm with $k \geq 2$ rather than based solely on the distance to a cluster.

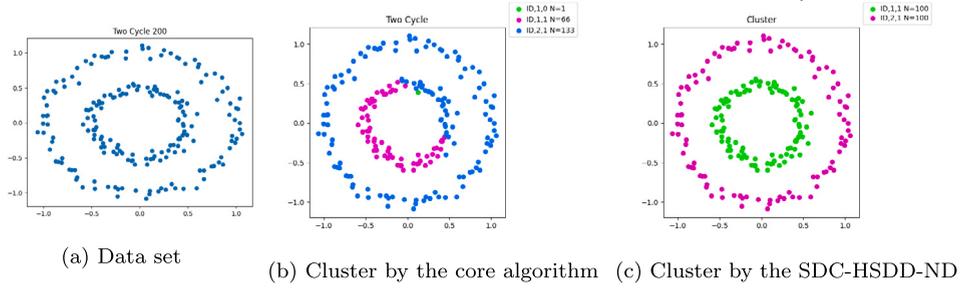

(a) Data set  (b) Cluster by the core algorithm  (c) Cluster by the SDC-HSDD-ND

**Fig. 4.** The dataset consists of two randomly sampled cycles, clearly forming two clusters: the inner one and the outer one. When clustering using the core algorithm, most points end up in clusters that are smaller than the minimum cluster size threshold and are thus merged into the nearest cluster that reaches the threshold.
Settings: $MaxIsoPointRho = 0.07$, $IsoNeiborK = 4$, $MinClusterPoint = 25$, $SearchNeiborK = 7$, $RhoCalculateK = 4$, $eps = 0.075$.

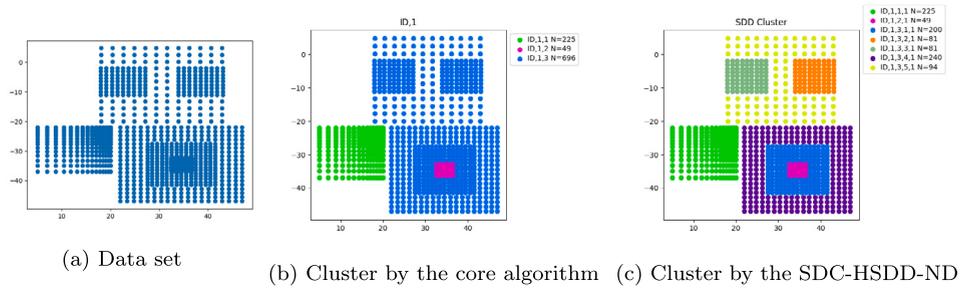

(a) Data set  (b) Cluster by the core algorithm  (c) Cluster by the SDC-HSDD-ND

**Fig. 5.** Extreme density: The region consists of 7 clusters integrated by the Triple-Square(Fig. 1), the Mountain-River(Fig. 2), and the Square (Fig. 3). However, the density in the top-right corner of the bottom-left square is extremely high. As a result, the normalized density of most regions (the blue cluster in (b)) is low (as it is much lower than the highest density). When clustering using the core algorithm, the blue region in (b) is considered to have a low secondary differential of densities. This occurs because, after normalization, the densities of points in it are extremely low, preventing it from being distinguished.
Settings: $MaxIsoPointRho = 0.07$, $MinClusterPoint = 25$, $IsoNeiborK = 4$, $SearchNeiborK = 7$, $RhoCalculateK = 4$, $eps = 0.075$.

### 3.6. Problems and solutions

There are two main issues when applying the core algorithm to clustering.

Firstly, the core algorithm might fail to cluster datasets where high-density clusters are separated by low-density regions that have cluster-level structures but lack coherent, refined structures within clusters, as shown in Fig. 4. The clustering result from simply applying the core algorithm is shown in (b) of Fig. 4, which fails to cluster the two cycles as in the $K$-mean algorithm (though it might be better). This happens because there are small clusters of sufficient size in both the inner and outer cycles, but most points are in small clusters and are therefore merged into the closest one as in the $K$-mean algorithm. Fortunately, this issue can be resolved using traditional density-based algorithms such as clustering by the $k$-NN version of DBSCAN described in Section 2, which is a special case of the core algorithm presented above. Therefore, an immediate solution could be to apply the algorithm with $mode = kNN$ for clustering via the $k$-NN clustering algorithm, followed by applying the algorithm with $mode = SD$, which implements the core algorithm presented above normally. The clustering result is provided at (c) of Fig. 4.

Another problem is that high-density points whose densities are low compared to the highest-density one might need to be separated into different clusters but may fail to do so, for example, please refer to (b) of Fig. 5. This issue might arise because the right-up density in the left-down square is much higher than most regions, such that the normalized densities in some regions (the blue region) are very low, leading to small secondary differentials and thus preventing proper classification. This problem is common in most non-hierarchical density-based clustering algorithms. The solution is to apply hierarchical clustering by repeating the core algorithm, and the result is provided in (c) of Fig. 5.

### 3.7. The SDC-HSDD-ND algorithm

After the investigations in the above subsection, the hierarchical algorithm, referred to as SDC-HSDD-ND, is ready to be presented. In summary, SDC-HSDD-ND is implemented as follows.

Step 1: Implement the $k$-NN clustering algorithm.

Step 2: Refine the $k$-NN clustering result using the core algorithm, namely SDC-SDD-ND without detecting isolated points, until no further refinements can be made.

Step 3: Run the core algorithm with isolated-point detecting on the result from Step 2 and obtain the final result.

The detailed pseudocode is proposed in Algorithm 2.

**Algorithm 2** The hierarchical algorithm: SDC-HSDD-ND.

**Input:** Items in the core algorithm
    %The minimal requirement only contains the data set
**Input:** $MidResult$
    %(Optional) Whether return middle results
**Input:** $KON$
    %(Optional) Detect isolated points or not
**Input:** $mode$
    %(Optional) $mode = SD$ represents implement the core algorithm with $eps$, $mode = kNN$ means implement the core algorithm with $eps = 4$, namely clustering
    by $SearchNeiborK$-NN
**Output:** Cluster result
1: $FinalCluster = []$
    %Collects all final clusters
2: Implement the core algorithm with $KON = False$ and $mode = kNN$
3: $CRefine = \{$All clusters$\}$
4: **while** $CRefine \; != \emptyset$ **do**
5:     $T =$Cluster result in implementing the core algorithm on $CRefine[0]$ with $mode = SD$ and $KON = False$
6:     Remove $CRefine[0]$ from $CRefine$
7:     **if** $|T| == 1$ **then**
8:         Implement the core algorithm on the cluster in $T$ with $mode = SD$ and $KON = True$, add results into $FinalCluster$
9:     **else**
10:        Add the clusters in $T$ into $CRefine$
11:     **end if**
12: **end while**
13: Return $FinalCluster$

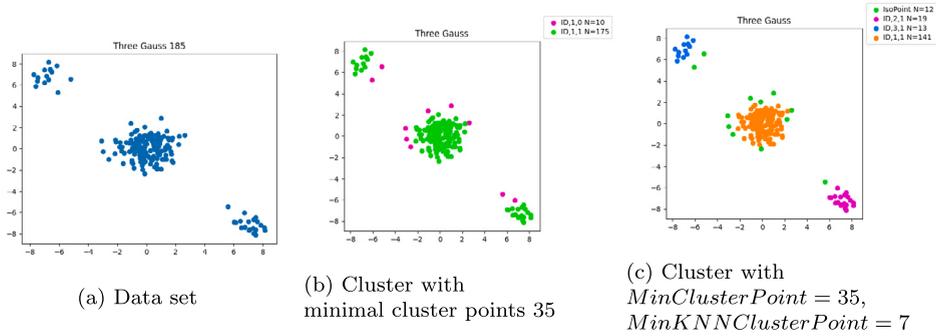

(a) Data set

(b) Cluster with
minimal cluster points 35

(c) Cluster with
$MinClusterPoint = 35$,
$MinKNNClusterPoint = 7$

**Fig. 6.** Three Gauss: The dataset consists of three random Gaussian sampled clusters but the ones in the corners have very few members (15 and 25, respectively). When clustering with $MinClusterPoint = 35$, the clusters in the corners are too small and thus are merged into the middle one as in (b). The issue can be solved by choosing another bound $MinKNNClusterPoint$ as the minimal cluster length threshold when $mode = KNN$ as in (c).
Settings: $MaxIsoPointRho = 0.07$, $MinClusterPoint = 35$, $IsoNeiborK = 4$, $SearchNeiborK = 7$, $RhoCalculateK = 4$, $eps = 0.075$.

### 3.8. The choices of $eps$ and the minimal length on merging

The last thing left without discussion in the above subsections is the choice of $eps$ (the threshold of secondary directed differential of points) and $MinClusterPoint$ (the minimal number of points required to form a cluster).

The choice of $MinClusterPoint$ depends on how small a cluster should be aborted. As the over-refinement problem is one of the main concerns in the algorithm, one should expect the threshold of a cluster to be high. Here, the default of $MinClusterPoint$ is set to 35 after several tests. On the other hand, a coefficient related to the number of data might be a better choice to be the threshold of a cluster since a larger dataset seems to have a higher threshold. Therefore, the final choice of the minimal length of a cluster could be $max(MinClusterPoint, (1 - f) \times N)$, where $0 \leq 1 - f < 1$ is a fixed fraction and $N$ is the number of data points.

However, a large $MinClusterPoint$ might cause another problem that small clusters could easily be absorbed, which could sometimes merge distant clusters. For instance, see (b) of Fig. 6, where the two clusters in the corner are too small with 15 and 20 members, respectively, and thus are merged into the center one if the minimal length of a cluster is set to be 35 simply. The issue might not be serious in $mode = SD$ since the refinements are implemented after clustering by neighbors, but it might be a problem in $mode = KNN$. A suggestion to reduce the problem is relaxing the condition of the minimal cluster, namely choosing another bound $MinKNNClusterPoint \leq MinClusterPoint$ as the minimal requirement for numbers of points in a cluster in $mode = KNN$. In such a mode, all clusters whose length is smaller than $MinKNNClusterPoint$ are merged into the cluster of isolated points, and the default of $MinKNNClusterPoint$ is chosen to be 7. The testing result is provided in (c) of Fig. 6.

On the other hand, the choice of $eps$ could be essential. In principle, it depends on how accurately one wants to detect structures. If $eps$ is set to be equal to or larger than 4, then the core algorithm is the one clustering simply by $SearchNeiborK$-NN without detecting structures. This demonstrates that clustering by $k$-NN is a special case of the above algorithm. Generally, the decrease of $eps$ represents a tighter restriction in clustering, where more accurate structures could be detected. However, the tighter condition also means that less accurate structures are more likely to be split into different clusters, which could increase the risk of over-refinement, especially in those clusters without regular structures. The default of $eps$ in the algorithm is set to be 0.075 after several tests.

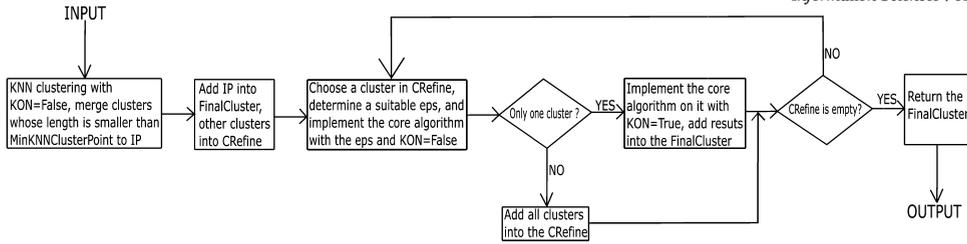

**Fig. 7.** The flow chart of the final algorithm.

However, the default setting on $eps$ might not be sufficient for certain tasks. A better way might be that $eps$ could self-adjust. Therefore, we suggest that the algorithm begin with a small $eps$ and then enlarge it until a suitable one is found. Despiteness, the initialization of $eps$ might still be a problem. It could neither be so small that leads to serious over-refinements nor be too large to detect the structures. Furthermore, the exploration of $eps$ could consume extra calculations, which should be taken into account.

Based on the discussions above, we suggest employing the clustering algorithm with self-adjust $eps$ chosen as follows. Whenever a dataset employing $mode = SD$ to cluster, a suitable $eps$ is chosen by running the clustering algorithm with $eps$, initialized by a small $Mineps$, recording the number of clusters whose length reaches $MinClusterPoint$, and then followed by another run with $eps = eps + adjust$, where $adjust$ is the chosen step in enlarging $eps$. If the number of effective clusters is reduced, then $eps - adjust$ could be considered as the suitable one, otherwise repeat the procedure until a suitable $eps$ is found or $eps \geq Maxeps$, where $Maxeps$ is the maximal choice of $eps$. In the default settings, $Mineps$ is set to 0.045, $Maxeps$ is set to 0.075, and $adjust$ is set to 0.005.

### 3.9. The final algorithm: SDC-HSDD-NDSA

Finally, the algorithm integrating all of the considerations is ready to generate, please see Fig. 7, and refer to the following steps as a summary.

Step 1 (Lines 3-9): Cluster with $k$-NN clustering algorithm without detecting isolated points and merge the clusters that are too small into a special isolated-point cluster.

Step 2 (Lines 10-13): For the remaining clusters, first deal with those that are small enough such that at most one cluster is contained, but not so small that consists of isolated points only, by implementing $k$-NN algorithm with isolated-point detecting, and adding the results to the final cluster result.

Step 3 (Line 14): Let the set of the remaining clusters, namely the set of big clusters in which more than one cluster might be contained, be $CRefine$.

Step 4 (Lines 15-31): Select a cluster in $CRefine$, and choose a threshold $eps$ of the secondary directed differential over which the core Algorithm 1 can provide the largest number of effective clusters, namely whose length is greater than a predefined threshold.

Step 5 (Line 32): Cluster the selecting cluster by the core algorithm with directed secondary differential threshold $eps$ without detecting isolated points.

Step 6 (Line 33): If the clustering result in step 5 consists of a unique effective cluster, recluster it with isolated-point detecting and adding the result into the final results. If not, add the result in step 5 into $CRefine$. In both cases, remove the selected cluster in step 4.

Step 7 (Line 34 (The *'while'* cycle end in Line 34)): Repeat Step 4 to Step 6 until $CRefine = \emptyset$ and get the final result.

Step 8 (Line 35): Redistribute the isolated points by a suitable scheme, for example, to the closest effective clusters if needed.

The detailed pseudocode is given in Algorithm 3, where the coefficient $IOC$ represents whether isolated points are required to be merged into a single cluster. If it is $True$, all isolated points would be merged into a single cluster, and otherwise, local isolated points would be displayed in different clusters.

## 4. Experimental results

In this section, experimental results are provided. The algorithm is tested on several datasets, including synthetic ones and real ones, and compared to previous unsupervised clustering algorithms. Furthermore, our algorithm is also implemented on noisy datasets to confirm its robustness against noises. More experimental results, including datasets combining different clusters with different granularities and random isolated points, are provided in the supplementary materials.

In the whole section as well as the remaining paper, $Noise = x$ represents that points are added with Gaussian noises with the standard deviation $\sigma = x \times d$, where $d$ is the granularity, namely $d_x = max(\{x|(x,y) \in Dataset\})\text{-}min(\{x|(x,y) \in Dataset\})$, $d_y = max(\{y|(x,y) \in Dataset\})\text{-}min(\{y|(x,y) \in Dataset\})$, and $d = max(d_x, d_y)$, for 2D data.

**Algorithm 3** The final algorithm with self-adaption: SDC-HSDD-NDSA.

---

**Input:** Items in the SDC-HSDD-ND algorithm
    %The minimal requirement only contains the data set
**Input:** $MinKNNClusterPoint$
    %(Optional) The minimal length of clusters in $mode = KNN$
**Input:** $IOC$
    %(Optional) If $IOC = True$, then all isolated points will be merged into one cluster. Otherwise, isolated points will be displayed locally.
**Input:** $Mineps$
    %(Optional)
**Input:** $Maxeps$
    %(Optional)
**Input:** $adjust$
    %(Optional)
**Input:** $Redistribute\_Isolated\_Clusters$
    %(Optional) If $Redistribute\_Isolated\_Clusters = True$, then the isolated points will be redistributed to the closest effective clusters.
**Output:** Cluster result
 1: $FinalCluster = []$
    %Collects all final clusters
 2: $TotalIP = []$
    % Valid only when $IOC = True$. Collect isolated points
 3: Implement the core algorithm with $KON = False$ and $mode = kNN$ without merging small clusters
 4: Merge all clusters whose length is smaller than $MinKNNClusterPoint$ into the isolated-point cluster
 5: **if** $IOC == True$ **then**
 6:     Merge the isolated-point cluster into $TotalIP$
 7: **else**
 8:     Add the isolated-point cluster into $FinalCluster$
 9: **end if**
10: **for** Each remaining cluster whose length is smaller than $MinClusterPoint$ **do**
11:     Implement the core algorithm with $mode = KNN$ and $KON = True$, merge the isolated-point cluster into $TotalIP$ if $IOC == True$
12:     Add the cluster result into $FinalCluster$
13: **end for**
14: $CRefine = \{$All remaining clusters$\}$
15: **while** $CRefine\ != \emptyset$ **do**
16:     $EPS = Mineps$
17:     $ClusterNumber = 0$
18:     **while** $adjust > 0$ **do**
19:         Implement the core algorithm without merging with $eps = EPS$, $mode = SD$ and $KON = False$ on $CRefine[0]$
20:         **if** The number cluster whose length reaches $MinClusterPoint$ is less than $ClusterNumber$ **then**
21:             $EPS = EPS - adjust$
22:             **break**
23:         **else**
24:             $ClusterNumber = $The number of clusters
25:             $EPS = EPS + adjust$
26:         **end if**
27:         **if** $EPS > Maxeps$ **then**
28:             $EPS = Maxeps$
29:             **break**
30:         **end if**
31:     **end while**
32:     $T = $Cluster result in implementing the core algorithm on $CRefine[0]$ with $mode = SD$, $eps = EPS$, and $KON = False$
33:     Implement steps 6 to 11 in the SDC-HSDD-ND algorithm but merge all isolated-point clusters into $TotalIP$ if $IOC = True$
34: **end while**
35: Add $TotalIP$ into $FinalCluster$ if $IOC = True$ and $TotalIP != \emptyset$, or redistribute points in the isolated cluster if need.
36: Return $FinalCluster$

---

### 4.1. Effectiveness on both synthetic and real datasets

The following experiments demonstrate the effectiveness of the SDC-HSDD-NDSA algorithm, where we compare our algorithm to previous clustering algorithms including DBSCAN, OPTICS, and Brich, on synthetic as well as real datasets. Experimental results demonstrate that our algorithm could achieve satisfactory results in some situations where previous algorithms do not perform well. In detail, Fig. 8 and Table 1 display the results on synthetic datasets with or without structures and Fig. 9 displays the results on real images of some flags downloaded from the internet. Please refer to the captions of the figures for implementation illustrations.

### 4.2. The need for hierarchy

The following result in Fig. 10 can demonstrate that only employing $k$-NN algorithm could be insufficient. We combine the clustering results by SDC-HSDD-NDSA and by simply employing $k$-NN clustering with different $k$. It shows that the simple employment of the $k$-NN clustering algorithm introduced in Section 2 fails in $k = 7$, and has invalid on Gaussian clusters (top-left) when $k$ decreases to 4 but still fails to cluster the TripleSquare (bottom-right). On the other hand, the SDC-HSDD-NDSA not only succeeds in clustering all clusters but can also detect isolated points.

**Table 1**
Numerical results, namely ARI(Adjusted Rand Index) and NMI(Normalized Mutual Information), on datasets of Fig. 8. a higher score represents a better prediction.

| ARI/NMI | Brich | DBSCAN | OPTICS | Ours |
|---------|-------|--------|--------|------|
| TC | 0.11/0.39 | 1.00/1.00 | 0.35/0.42 | 1.00/1.00 |
| TL | 0.74/0.81 | 0.57/0.73 | 0.55/0.66 | 1.00/1.00 |
| TG | 0.25/0.39 | 0.12/0.21 | 0.24/0.42 | 0.65/0.79 |
| SQ | 0.00/0.00 | 0.00/0.00 | 1.00/1.00 | 1.00/1.00 |
| MR | 0.17/0.45 | 0.00/0.00 | 0.59/0.64 | 1.00/1.00 |
| IG | 1.00/1.00 | 1.00/1.00 | 0.65/0.78 | 1.00/1.00 |
| CG | 0.15/0.37 | 0.03/0.01 | 0.75/0.73 | 0.93/0.87 |
| SDD | 0.32/0.60 | 0.00/0.00 | 0.26/0.54 | 1.00/1.00 |
| Average | 0.34/0.50 | 0.34/0.37 | 0.55/0.65 | 0.95/0.96 |

### 4.3. Robustness

The results of the SDC-HSDD-NDSA run on the datasets in the introduction are provided in Fig. 11, Fig. 12, and Fig. 13, which demonstrate that the algorithm is effective on these datasets and robust over noises.

## 5. Discussions

### 5.1. Asymmetry

The SDC-HSDD-NDSA algorithm is asymmetric, i.e., choosing different start points in clustering might provide different results. To allow a unique clustering result, the algorithm is set to choose the unclassified point with the highest density as the start point when forming a new cluster.

### 5.2. Detecting isolated points

As shown in the above sections, the SDC-HSDD-NDSA algorithm can detect isolated points. The isolated clusters consist of points that are not distributed to a cluster whose length reaches the threshold $MinKNNClusterPoint$ in $mode = KNN$ and those with normalized density lower than the threshold $MaxIsoPointRho$ in $mode = SD$. They are either merged into a single cluster (for $IOC = True$) or displayed separately on $mode = KNN$ and $mode = SD$ with the latter further separated locally into different isolated clusters (for $IOC = False$).

### 5.3. Complexity

The time complexity of the SDC-HSDD-NDSA algorithm can be discussed as follows, assuming the worst case unless otherwise pointed out and the dimension of the data is $d$.

(1) Calculate $k$-NN: $O(d \times N^2)$ for the worst case and $O(d \times N \times log N)$ on average by the KD-tree method.

(2) Calculate density and normalization: $O(N)$.

(3) Calculate the density-differentials of points in $k$-NN of each point: $O(N)$.

(4) Determine isolated points $IP$: $O(N)$.

(5) Cluster: $O(N^2)$ for starting with highest-density points. Determining the starting point with the maximal density among all non-clustered points of each new cluster is the bottleneck, in which the time complexity could be $O(N^2)$ in the worst case that every cluster consists of exactly one data and the dataset is ordered by the increasing order of the density. The time complexity for absorbing points is $O(|C_i|)$ for the cluster $C_i$ and thus $O(N)$ in total since $\sum_i |C_i| = N$. Therefore, the final complexity is $O(N^2)$ for the worst case. However, the procedure can be accelerated to $O(N)$ if new clusters are started by an arbitrarily chosen non-clustered point instead of the ones with the highest density.

(6) Merge: $O(d \times N^2)$ in the worst case and $O(d \times N \times log N)$ on average by the KD-tree method. The merge procedure is essentially a $k$-NN classifier with $k = 1$, where data in a cluster whose length reaches the minimal requirement is labeled with its cluster, and the clusters of the redistributed data are predicted by the $k$-NN classifier.

Therefore, the core algorithm could run with the time complexity $O(d \times N^2)$ in the worst case, and $O(d \times N \times log N)$ on average by the KD-tree method if one does not require the clusters starting by the non-clustered points with the highest-density. The self-adjust procedure could be considered as repeating the core algorithm a few (but not too many) times, and thus would not increase the time complexity. Finally, in the worst case, namely the hierarchical cluster-tree is as unbalanced as possible, there could be $t + 1$ hierarchies with $t \approx log_f(\frac{MinClusterPoint}{N})$, resulting in the total time complexity $\sum_{i=0}^{t} g(f^i N)$, where $N$ is the number of data, $1 - f$ is the fixed fraction stated in Section 3.8, and $g(x)$ denotes the time complexity of the (non-hierarchical) self-adaption algorithm in $x$ data. Hence, the time complexity of the SDC-HSDD-NDSA algorithm could be $O(d \times N^2)$ in the worst case that $g(x) = O(d \times x^2)$, while $\sum_{i=0}^{t} O(d \times f^i N \times log f^i N) = O(d \times N \times log N)$ in the average case that $g(x) = d \times x \times log x$, by choosing $f$ strictly small than 1, since equation (1).

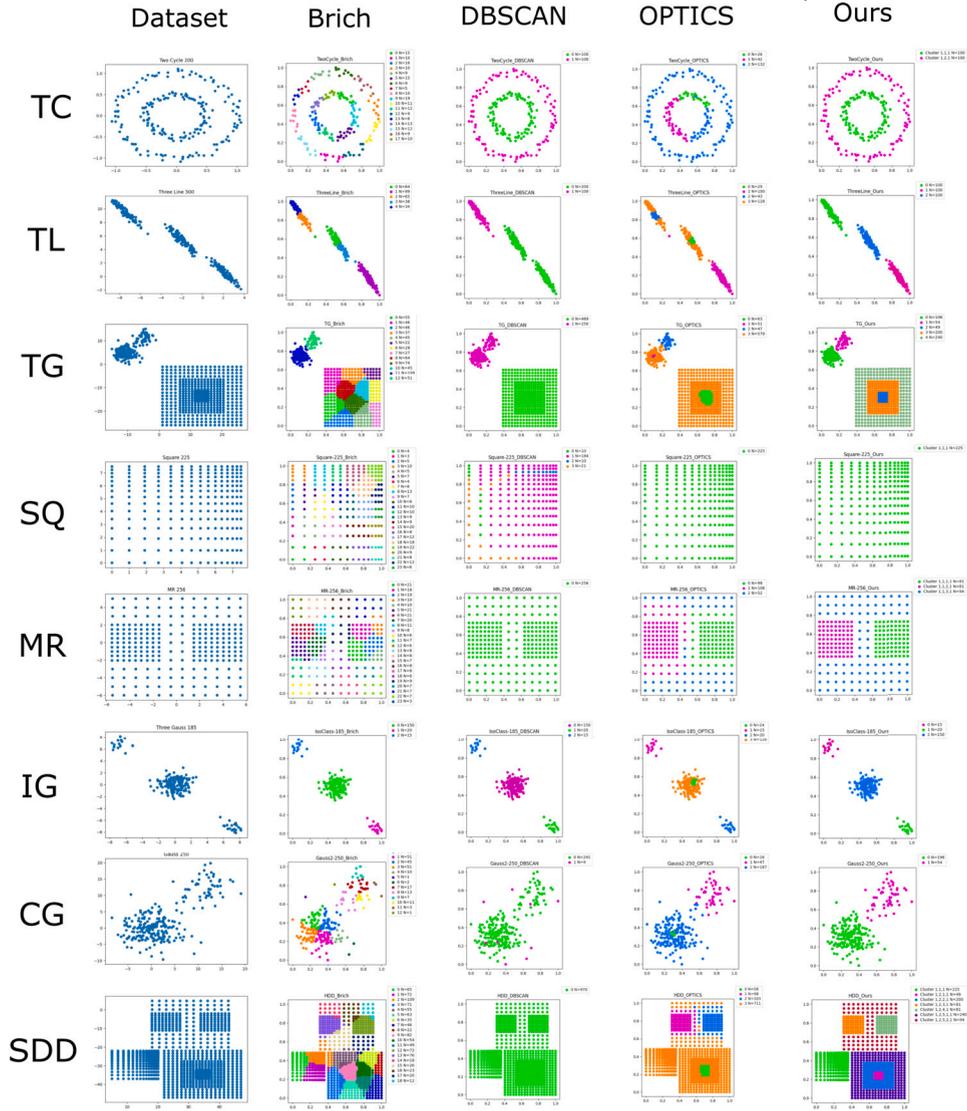

**Fig. 8.** Clustering results on 8 synthetic datasets. Those datasets are representative with different properties explained below, respectively. They represent problematic situations in which previous algorithms could fail. On the other hand, one can find that our algorithm could work well in these cases.

The first row (Two cycles, TC) is the dataset consisting of two cycles obtained by randomly sampling. The second row (Three lines, TL) is the dataset consisting of three lines obtained by randomly sampling. The third row (Triple square with Gaussian samples, TG) is the dataset consisting of two random Gaussian sampling clusters on the top-left and three clusters with different densities but which are not separated by the low-density areas on the bottom-right. The fourth row (Square, SQ) is the dataset consisting of a square whose density changes gradually. The fifth row (MR) is the dataset consisting of two higher-density clusters separated by a lower-density one, but the low-density area also forms a cluster. The sixth row (IG) is the dataset consisting of three clusters obtained by randomly sampling Gaussian data but two of them are small. The seventh row (CG) is the dataset consisting of two clusters obtained by randomly Gaussian sampling but are close to each other. The last row (SDD) is the dataset consisting of seven clusters obtained by combining some datasets above with different granularities and they are close to each other.

In OPTICS, the *minimal_samples* is chosen to be 15, and in Brich the *threshold* is chosen to be 0.1, while in DBSCAN *eps* is chosen to be 0.13, to obtain better results. In our algorithm, we follow the default setting mentioned in Section 3 and redistribute isolated points to the closest effective clusters. Moreover, all datasets are preprocessed by the same Max-Min scheme to rescale the data to $(0, 1)$ on every axis, respectively.

All the algorithms use the same coefficients over all the above datasets since the generality of the coefficients is somehow expected. However, the coefficients are selected to obtain better results. The numerical results for the algorithms over the above datasets are displayed in Table 1.

$$d \times N \times log N \leq \sum_{i=0}^{t} d \times f^i N \times log f^i N \leq \sum_{i=0}^{t} d \times f^i N \times log N \qquad (1)$$

$$\leq d \times N \times log N \times \frac{1 - f^{t+1}}{1 - f} \leq d \times N \times log N \times \frac{1}{1 - f}$$

A similar argument can show that the space complexity of the SDC-HSDD-NDSA algorithm could be the same as a $k$-NN searching algorithm, since the space complexity of procedures is not larger than $O(N)$ except the searching of $k$-NN, which is not less than $O(N)$.

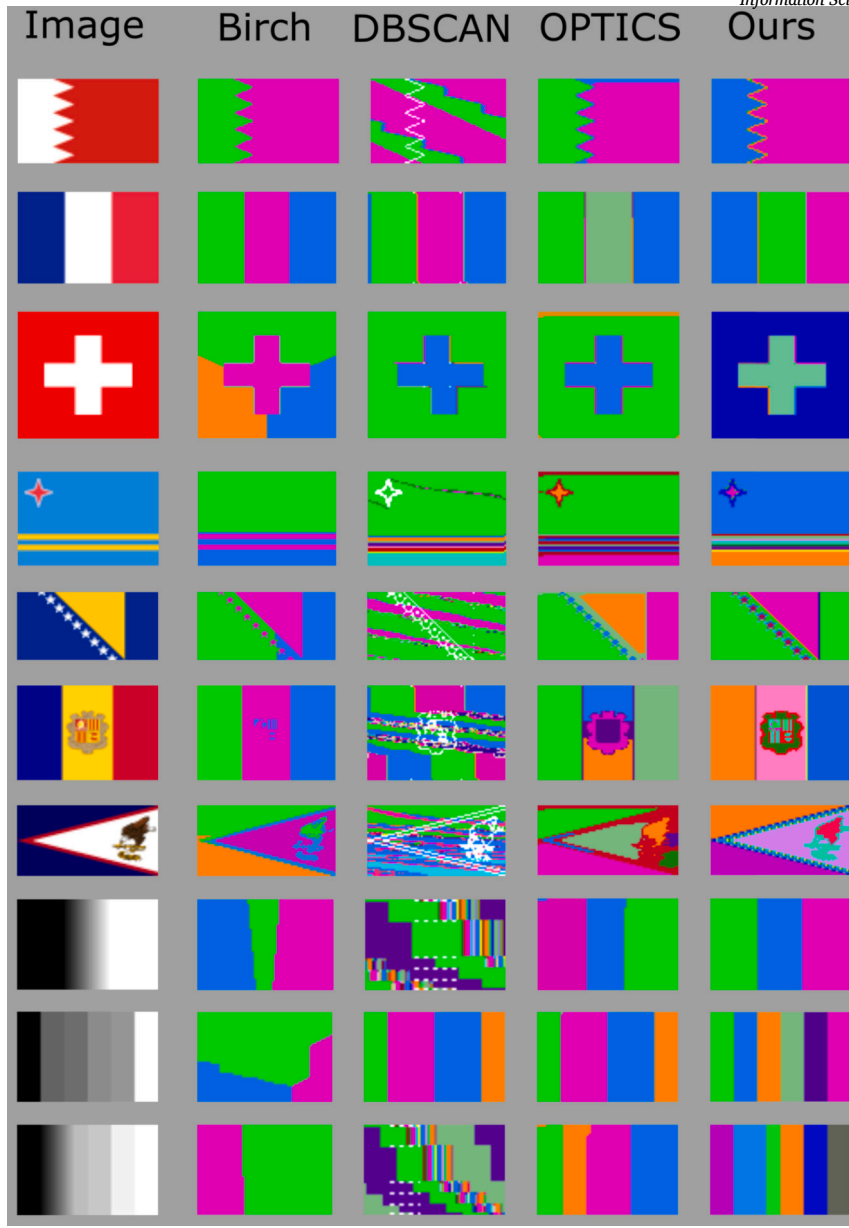

**Fig. 9.** Clustering results on real datasets consisting of flags, where the 2D coordinates together with the RGB colors (for the first seven rows) or the grayscale (for the last three rows) of pixels form $5D$ or $3D$ vectors and are employed as the features of pixels of an image. Each feature component of data is divided by the maximal value, respectively in the component to rescale it to $(0,1)$. The last three rows are synthetic and serve as color cards. The third row from the bottom consists of three clusters, the left area is totally black, the middle area is changed gradually from black to white, and the right area is totally white. The second row from the bottom consists of six clusters, the most-left area is totally black and is significantly different from the second-left area, the second-left area is a little bit different from the third-left area, similarly the third-right area is a little bit different from the second-right area, while the most-right area is totally white and significantly different from the others. The last row also consists of six clusters and is a combination of the above cases, the most-left area is totally black, followed by a gradually changed area (second-left), and two groups (the fourth-right and the third-right clusters form a group and the second-right and the most right clusters form another) of little-changed areas with each consisting of two clusters.

In Brich and OPTICS, we follow their default settings. In DBSCAN, we choose $eps = 0.05$. In our algorithm, we follow the default setting except for $SearchNeiborK = 9$, $RhoCalculateK = 7$, $MinKNNClusterPoint = 13$ and $eps = 0.18$ without self-adaption, with isolated points not being redistributed.

Our algorithm could outperform previous ones in the real dataset as well as in the color cards.

### 5.4. Extendable

The differential in higher levels as well as the combination of differentials in different levels might also be employed in clustering. Also, the normalizing scheme could be applied to the differentials of densities, not only for the densities. However, the employment methods and the effectiveness of them might need another study.

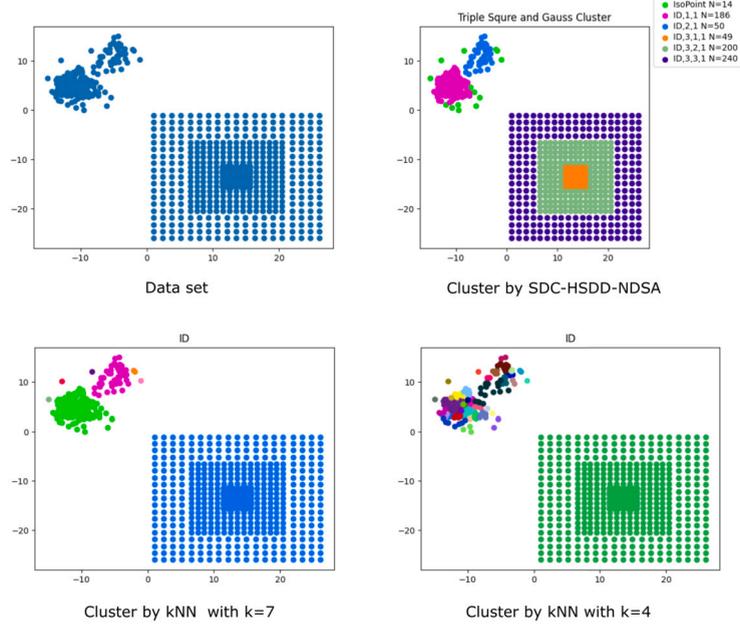

**Fig. 10.** Clustering results of the dataset combining the Triple-Square in Fig. 1 and two Gaussian samples. Our algorithm successfully clusters the data and also detects the isolated points, while the *k*-NN algorithm clusters the Gaussian samples but fails in the Triple-Square when $k = 7$ and still fails when $k$ is dropped to $k = 4$ in which even the Gaussian samples cannot be successfully clustered.

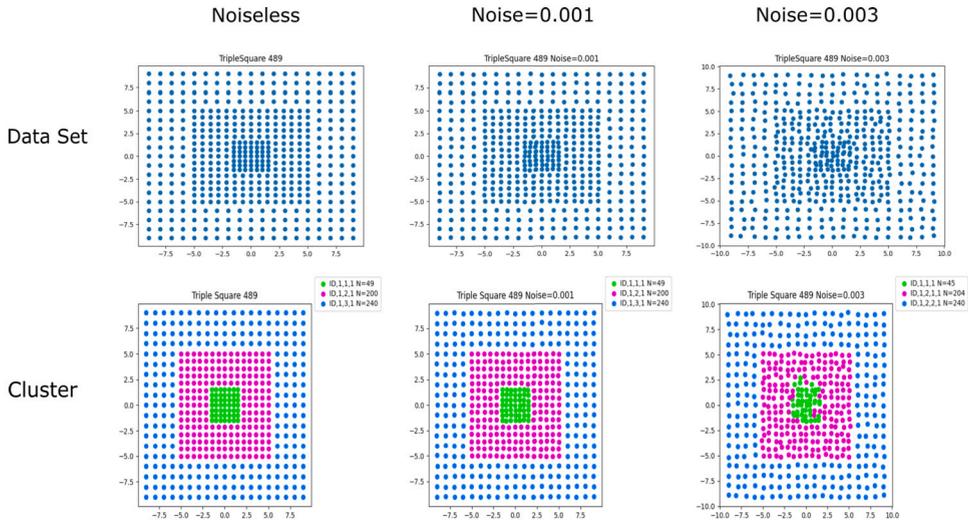

**Fig. 11.** Clustering results of the dataset TripleSquare in Fig. 1 with random Gaussian noises, where $Noise = x$ represents the standard deviation of the Gaussian noises being set to $x$. Results show that our algorithm is still effective when $Noise = 0.003$. The ARI/NMI are 1.00/1.00, 1.00/1.00, 0.95/0.90, for noiseless, $Noise = 0.001$, $Noise = 0.003$, respectively.

## 5.5. A review of the addressed problems and solutions

The problem addressed and the technique employed by our algorithm are mainly in four aspects.

Firstly, previous algorithms fail to detect structures in the clusters, which is caused by the simple employment of density threshold. Although the efforts in calculating the density of data by varied schemes do improve the accuracy of clustering, they cannot detect the structures in high-density regions such as the cases shown in Fig. 1, since even theoretically, no suitable simple density threshold can be found. The problem is essentially attributed to the misunderstanding that the border of clusters is characterized by the absolute density, while indeed the borders are characterized by the differential of densities. Therefore, we consider the differential of densities as the cluster criterion rather than the density itself, which addresses the problem.

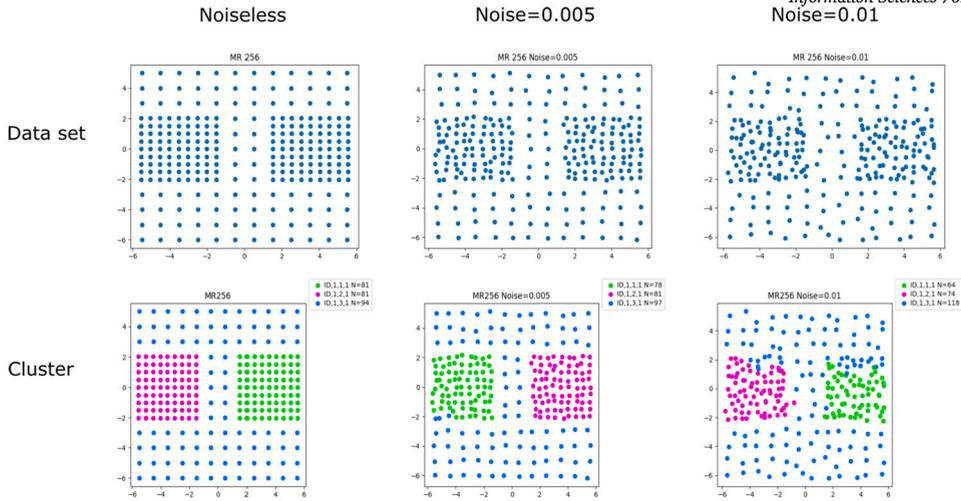

**Fig. 12.** Clustering results of the dataset MR in Fig. 2 with random Gaussian noises. Results show that our algorithm is still effective when $Noise = 0.01$. The ARI/NMI are 1.00/1.00, 0.96/0.95, 0.71/0.72, for noiseless, $Noise = 0.005$, $Noise = 0.01$, respectively.

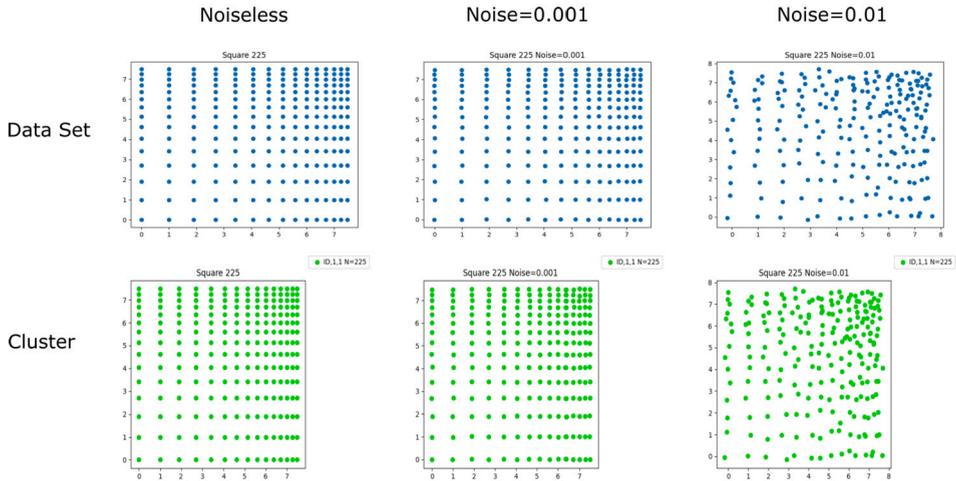

**Fig. 13.** Clustering results of the dataset Square in Fig. 3 with random Gaussian noises. Results show that our algorithm is still effective when $Noise = 0.01$. The ARI/NMI are 1.00/1.00, 1.00/1.00, 1.00/1.00, for noiseless, $Noise = 0.001$, $Noise = 0.01$, respectively.

Secondly, even employing the first differential of densities could be insufficient for structure detecting since there can be structures of a single cluster with a higher first differential while there can also be structures of different clusters with lower differential densities, such as in Fig. 1 and Fig. 3. Therefore, the secondary differential might be more effective.

Thirdly, the differential of data with dimension at least 2 is essentially directed, and thus employing features of single points such as gradient value could result in error clustering, such as the case of Fig. 2.[5]

Lastly, the ability of granularity independence is expected in most clustering algorithms but is not trivial to obtain. Normalization might be a solution but could be insufficient since it is only suitable for the situation where clusters have similar granularity, and might result in the missing or merging of clusters if the highest density in the dataset is so high that some effective clusters are considered as consisting of isolated data or are merged as a single cluster since they only consist of data with a similar (very low) normalized density, such as in Fig. 5 where some high-density regions have low normalizing densities (compared with the highest-density region).[6] To address this problem, we employ both hierarchy and normalizing methods, where the hierarchy scheme gradually provides similar granularity in hierarchy subsets, allowing normalizing schemes to be applied.

---

[5] The gradient value of the first two and last two rows are low while the gradient value of the middle columns is high. Therefore, when clustering by gradient values, the top two rows and the bottom two rows will be clustered into different clusters from each other and also from the middle two columns, which is counter-intuitive.

[6] Even normalizing by the average density might not address the issue since effective clusters might occur in a high-density region but whose density is low compared with the average density.

### 5.6. Limitations and future works

Despite the effectiveness in addressing the structure-detecting issues in clustering algorithms, the presented scheme might still be imperfect. Firstly as mentioned previously, different results might be obtained by different starting points. Secondly, the algorithm can still provide some errors in clustering, especially in noisy datasets. Thirdly, the self-adaptation coefficient calculated automatically by the algorithm might not be the optimal one in some cases.[7] Fourthly, even with the hierarchical clustering, the algorithm might still miss some non-obvious structures. Finally, the clustering algorithm tends to cluster all structures, still, missing their semantics.[8]

There are five main aspects for future work. Firstly, the structure-detecting problem is at the starting stage and the first model is just presented, thus more discussions and designs might be needed, for example in more general scenarios. Secondly, the presented model employs secondary differential, but higher-order differentials might also be valid. However, in what situations they are needed and how to use them remain problems. Thirdly, although the effectiveness of the presented model has been demonstrated, the acceleration of it might need another study. Fourthly, while some criteria such as ARI and NMI can be employed in measuring the capabilities of structure-detecting models, some others such as traditional cluster distance criteria are invalid since two structure clusters might usually be close to each other. Therefore, investigating the novel criteria for structure-detecting problems could be significant. Finally, previous datasets for clustering tasks are mostly without internal structures that need to be grouped into different clusters in high-density regions, and we have proposed some datasets with such structures. Nevertheless, more effective structure datasets might be needed for further research.

### 6. Conclusion

In conclusion, we have presented a novel density-based clustering algorithm, dubbed SDC-HSDD-NDSA, which can not only detect clusters separated by low-density regions as the previous ones can, but also detect structures in regions not separated by low-density regions, on which the previous ones are demonstrated impossible to be success. Therefore, it addresses the structure-detecting problem which has not been solved, and is even theoretically unsolvable by the previous density-based algorithms, extending the available range of clustering. The algorithm is convenient to employ since the minimal input only contains the dataset, while its complexity is the same as a $k$-NN searching algorithm. Meanwhile, some representative datasets problematic for previous algorithms have been provided, which could serve as materials for further research on structure clustering. Experimental results have demonstrated the effectiveness, robustness, and competitiveness of our algorithm. Also, the properties, complexity, extensibility, and limitations of the algorithm with future works have been discussed.

### CRediT authorship contribution statement

**Hao Shu:** Writing – review & editing, Writing – original draft, Visualization, Validation, Software, Resources, Methodology, Investigation, Formal analysis, Data curation, Conceptualization.

### Declaration of competing interest

The authors declare that they have no known competing financial interests or personal relationships that could have appeared to influence the work reported in this paper.

### Appendix A. Supplementary material

Supplementary material related to this article can be found online at https://doi.org/10.1016/j.ins.2025.121916.

### Data availability

Data is available GitHub with the address displayed in the article.

---

[7] In fact, we chose different coefficients in synthetic datasets and real datasets in experiments of Section 4.

[8] For example, In clustering the French flags, namely the second row in Fig. 9, our algorithm provides 5 clusters since there are indeed 5 colors one by one in the image, the blue areas, the area from blue to white (generated by down-sampling of the image), the white area, the area from white to red (also generating by down-sampling), and the red area. Although it is hardly viewed as an error, one might tend to ignore the down-sampling areas and view the image as containing only 3 clusters, with semantics. This issue can be addressed by increasing the threshold of a cluster and merging isolated points into effective non-isolated clusters. However, it might still be a problem if the image does contain small clusters that are not separated by low-density regions.